\documentclass[runningheads]{llncs}
\usepackage{graphicx}

\usepackage{tikz}
\usepackage{comment}
\usepackage{amsmath,amssymb} 
\usepackage{color}
\usepackage{multirow,multicol}

\usepackage{pifont}
\newcommand{\cmark}{\ding{51}}
\newcommand{\xmark}{\ding{55}}

\usepackage[accsupp]{axessibility}  

\usepackage{cancel}

\usepackage{hyperref}

\begin{document}
\pagestyle{headings}
\mainmatter
\def\ECCVSubNumber{6838}  

\title{DenseHybrid: Hybrid Anomaly Detection \\for Dense Open-set Recognition} 


\titlerunning{Hybrid Anomaly Detection for Dense Open-set Recognition}
%
\author{Matej Grcić \and
Petra Bevandić \and
Siniša Šegvić}
\authorrunning{M. Grcić et al.}
%
\institute{University of Zagreb \\ Faculty of Electrical Engineering and Computing \\ Unska 3, 10000 Zagreb, Croatia \\
\email{\{matej.grcic,petra.bevandic,sinisa.segvic\}@fer.hr}}
\maketitle
\begin{abstract}
Anomaly detection can be conceived
either through generative modelling
of regular training data
or by discriminating with respect
to negative training data.
These two approaches 
exhibit different failure modes.
Consequently, hybrid algorithms present 
an attractive research goal.
Unfortunately, dense anomaly detection requires
translational equivariance and
very large input resolutions.
These requirements disqualify
all previous hybrid approaches to the best of our knowledge.
We therefore design a novel hybrid algorithm
based on reinterpreting discriminative logits
as a logarithm of the unnormalized joint distribution $ \hat{p}(\mathbf{x},\mathbf{y})$.
Our model builds on a shared convolutional representation
from which we recover three dense predictions:
i) the closed-set class posterior $P(\mathbf{y}|\mathbf{x})$,
ii) the dataset posterior $P(d_{in}|\mathbf{x})$,
iii) unnormalized data likelihood $\hat{p}(\mathbf{x})$.
The latter two predictions are trained
both on the standard training data
and on a generic negative dataset.
We blend these two predictions
into a hybrid anomaly score
which allows dense open-set recognition
on large natural images.
We carefully design a custom loss
for the data likelihood
in order to avoid backpropagation
through the untractable
normalizing constant $Z(\theta)$.
Experiments evaluate our contributions
on standard dense anomaly detection benchmarks
as well as in terms of open-mIoU -
a novel metric for dense open-set performance. 
Our submissions achieve state-of-the-art
performance despite neglectable computational overhead
over the standard semantic segmentation baseline.
Official implementation: \url{https://github.com/matejgrcic/DenseHybrid}
\end{abstract}

\section{Introduction}
High accuracy, fast inference and small memory footprint of modern neural networks steadily expand the horizon of downstream applications.
Many exciting applications require advanced image understanding functionality provided by semantic segmentation \cite{farabet13pami}.
These models associate each pixel with a class from a predefined taxonomy.
They can accurately segment two megapixel images in real-time on low-power embedded hardware \cite{chao19iccv,orsic21pr,hong21arxiv}.
However, the standard training procedures assume the closed-world setup which may raise serious safety issues 
in real-world deployments. 
For example, if a segmentation model missclassifies an unknown object (e.g. lost cargo) as road, the autonomous car may experience a serious accident.
Such hazards can be alleviated by complementing semantic segmentation with dense anomaly detection.
The resulting dense open-set recognition models are more suitable for real-world applications due to ability to decline the decision in anomalous pixels.

Previous approaches for dense anomaly detection either use a generative or a discriminative perspective.
Generative approaches are based on density estimation \cite{blum19iccvw} or image resynthesis \cite{lis19iccv,biase21cvpr}.
Discriminative approaches use classification confidence \cite{hendrycks19arxiv}, a binary classifier  \cite{bevandic19gcpr} or Bayesian inference \cite{kendall17nips}.
These two perspectives exhibit different failure modes.
Generative detectors inaccurately disperse 
the probability volume \cite{nalisnick19iclr,serra20iclr,lucas19nips,zhang21icml} or rely on risky image resynthesis.
On the other hand, discriminative detectors assume training on full span of the input space, even including unknown unknowns \cite{hendrycks19iclr}.

In this work we combine the two perspectives into a hybrid anomaly detector.
The proposed approach complements a standard semantic segmentation model with two additional predictions:
i) unnormalized dense data likelihood $\hat{p}(\mathbf{x})$ \cite{blum19iccvw}, and ii)
dense data posterior $P(d_{in}|\mathbf{x})$ \cite{bevandic19gcpr}.
Both predictions require training with negative data \cite{hendrycks19iclr,bevandic19gcpr,biase21cvpr,chan21iccv}.
Joining these two outputs yields an accurate yet efficient dense anomaly detector which we refer to as DenseHybrid.

We summarize our contributions as follows.
We propose the first hybrid anomaly detector which allows end-to-end training and operates at pixel level.
Our approach combines likelihood evaluation and discrimination with respect to an off-the-shelf negative dataset.
Our experiments reveal accurate anomaly detection despite minimal computational overhead.
We complement semantic segmentation with DenseHybrid to achieve dense open-set recognition.
We report state-of-the-art dense open-set recognition performance according to a novel performance metric which we refer to as \textit{open-mIoU}.

\begin{figure}
    \centering
    \includegraphics[width=\linewidth]{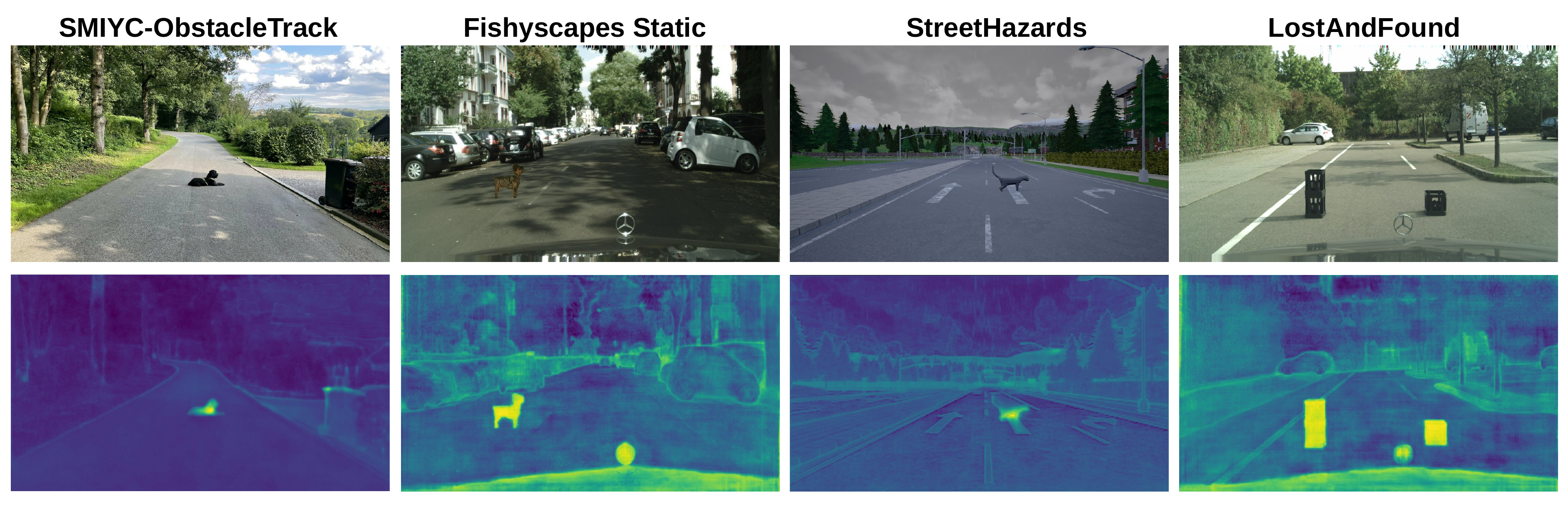}
    \caption{Qualitative performance of the proposed DenseHybrid approach on standard datasets. Top: input images. Bottom: dense maps of the proposed anomaly score}
    \label{fig:ood_detector}
\end{figure}

\section{Related Work}

Detecting samples which deviate from the generative process of the training data is a decades old problem \cite{hawkins80book}.
In the machine learning community this task is also known as anomaly detection or out-of-distribution (OOD) detection \cite{hendrycks17iclr}.
Early image-wide approaches
utilize max-softmax probability \cite{hendrycks17iclr}, input perturbations \cite{liang18iclr} ensembling \cite{lakshminarayanan17nips} or Bayesian uncertainty \cite{mukhoti18arxiv}.
More encouraging performance has been reported
by discriminative training against a broad negative dataset \cite{dhamija18nips,hendrycks19iclr,bevandic19gcpr,liu20neurips} or an appropriately trained generative model \cite{lee18iclr,grcic21visapp,zhao21arxiv}.

Another line of work detects anomalies by estimating the likelihood with a generative model.
Surprisingly, this research revealed that anomalies may give rise to higher likelihood than inliers \cite{nalisnick19iclr,serra20iclr,zhang21icml}.
Further works suggest that better performance
can be hoped for group-wise anomaly detection \cite{jiang22iclr}, however, this case has less practical importance.
Generative models can be encouraged to assign low likelihood in negative training data \cite{hendrycks19iclr}.
This practice may mitigate sub-optimal dispersion of the probability volume \cite{lucas19nips}.

Image-wide anomaly detection approaches can be adapted for dense prediction with variable success.
None of the existing generative approaches can deliver dense likelihood estimates.
On the other hand, concepts such as max-softmax and discriminative training with negative data are easily ported to dense prediction.
Many dense anomaly detectors are trained on mixed-content images obtained by pasting negatives (e.g. ImageNet, COCO, ADE20k) over regular training images \cite{bevandic19gcpr,chan21iccv,biase21cvpr}.
Discriminative anomaly detections may be produced by a dedicated OOD head which shares features  with the standard classification head.
Shared features improve OOD performance and incur neglectable computational overhead with respect to the baseline semantic segmentation model \cite{bevandic19gcpr}.
Recent approach \cite{chan21iccv} encourages large softmax entropy in negative pixels.

Anomalies can also be recognized in feature space \cite{blum19iccvw}.
However, this approach complicates the detection of small objects due to subsampled feature represenations and feature collapse \cite{lucas19nips,amersfoort21arxiv}.
Orthogonally to previous approaches, anomaly detector can be implemented according to dissimilarity between the input and a resynthesised image \cite{lis19iccv,biase21cvpr,vojir21iccv}.
The resynthesis is performed by a generative model conditioned on the predicted labels.
However, this approach is suitable only for uniform backgrounds such as roads \cite{lis19iccv}. 
Furthermore, it adds significant computational overhead making it inapplicable for real-time applications.

Our approach to dense anomaly detection is a hybrid combination of  discriminative detection and likelihood evaluation. 
Discriminative OOD detection has been introduced in \cite{bevandic19gcpr,hendrycks19iclr,dhamija18nips}. 
Contrary to all these approaches, we improve discriminative OOD detection through synergy with likelihood testing.
Dense likelihood evaluation has been accomplished by fitting a generative model
to discriminative features \cite{blum19iccvw}.
However, their approach is vulnerable to feature collapse\cite{lucas19nips,amersfoort21arxiv} due to two-phase training.
Moreover, detection of small outliers is jeopardized due to subsampling.
Contrary to their approach, our  method allows joint training with the standard dense prediction model and anomaly detection at full resolution.

We perform dense likelihood evaluation by reinterpreting logits as 
unnormalized joint likelihood \cite{grathwohl20iclr}. 
However, the method \cite{grathwohl20iclr} is completely unsuitable for dense prediction due to intractability of Langevin sampling at large resolutions.
We reformulate their method in order to allow training on mixed-content images and show that such adaptation dramatically simplifies the training by precluding backpropagation through intractable normalizing constant $Z(\theta)$.
To the best of our knowledge, the proposed design offers the first approach for dense likelihood evaluation that is suitable for end-to-end training.


We build an open-set recognition model by thresholding our hybrid anomaly score and combining it with the standard semantic segmentation predictions \cite{boult19aaai}.
The resulting model is suitable for simultaneous anomaly detection and recognition of inlier scenery.
We note that standard metrics for dense recognition performance \cite{everingham15ijcv} do not take into account the accuracy in anomalous samples.
This is not surprising since outlier pixels have been introduced only in recent dense prediction benchmarks \cite{zendel18eccv,blum21ijcv,chan21arxiv}.
Also, previous work on discrimination in presence of anomalous pixels was more focused on robustness of algorithms rather than on recognition performance \cite{zendel18eccv}.
Hence, we propose a novel anomaly-aware metric (open-mIoU) which measures the prediction quality both in inliers and the outliers, similarly to previous image-wide metrics \cite{sokolova09ipm,scherreik16taes}.

\section{Dense Recognition with Hybrid Anomaly Detector}

We propose a hybrid algorithm for dense anomaly detection based on unnormalized data likelihood and dataset posterior (Sec.\ \ref{sec:prob_view}).
The proposed hybrid anomaly detector extends the standard dense classifier to form dense open-set recognition model (Sec.\ \ref{sec:ratio_train}).
The resulting recognition model trains on mixed content images.


\subsection{Hybrid Anomaly Detection for Dense Prediction
\label{sec:prob_view}}

We represent RGB images with a random variable $\underline{\mathbf{x}}$.
Variable $\underline{\mathbf{y}}$ denotes the corresponding pixel-level predictions, while the binary random variable $\underline{d}$ models whether a given pixel belongs to the inliers or outliers. 
We denote a realization of a random variable without the underline.
Thus, $P(\mathbf{y}|\mathbf{x})$ is a shortcut for $P(\underline{\mathbf{y}}=\mathbf{y}|\underline{\mathbf{x}}=\mathbf{x})$.
We write $d_{in}$ for inliers and $d_{out}$ for outliers.
Thus, $P(d_{in}|\mathbf{x})$ denotes a dense posterior probability that a given pixel is an inlier \cite{hendrycks19iclr,bevandic19gcpr}.
Conversely, $p(\textbf{x})$ denotes dense likelihoods of patches centered at a given pixel.

We build upon reinterpretation of logits $\mathbf{s}$ produced by a discriminative model $P(\mathbf{y}|\mathbf{x})=\mathrm{softmax}(f_{\theta_2}(q_{\theta_1}(\mathbf{x})))$ \cite{grathwohl20iclr}.
We reinterpret the logits as unnormalized joint log-density of input and labels:
\begin{equation}
\label{eq:joint}
    p(\mathbf{y}, \mathbf{x}) = \frac{1}{Z} \hat{p}(\mathbf{y}, \mathbf{x}) :=  \frac{1}{Z} \exp{\mathbf{s}}, \quad \,  \mathbf{s} = f_{\theta_2}(q_{\theta_1}(\mathbf{x})).
\end{equation}
Note that $q_{\theta_1}$ produces pre-logits $\mathbf{t}$ based on which $f_{\theta_2}$ computes logits $\mathbf{s}$.
Hence, $q_{\theta_1}$ and $f_{\theta_2}$ form the standard discriminative model.
$\hat{p}(\mathbf{y}, \mathbf{x})$ denotes unnormalized joint density across data $\underline{\mathbf{x}}$ and labels $\underline{\mathbf{y}}$, while $Z$ denotes the corresponding normalization constant.
As usual, computing $Z$ is intractable since it requires evaluating the unnormalized distribution for all realizations of $\underline{\mathbf{y}}$ and $\underline{\mathbf{x}}$.
Throughout this work we conveniently eschew the evaluation of $Z$ in order to enable efficient training and inference.

Standard discriminative predictions 
are 
easily obtained through Bayes rule:
\begin{equation}
   P(\mathbf{y}|\mathbf{x}) = \frac{p(\mathbf{y},\mathbf{x})}{\sum_{\textbf{y}} p(\mathbf{y},\mathbf{x})} = \frac{\exp{\mathbf{s}}}{\sum_i \exp{\mathbf{s}_i}} = \mathrm{softmax}(\mathbf{s}).
\end{equation}
Hence, we can recover the unnormalized joint density (\ref{eq:joint}) through the standard closed-world discriminative learning over K classes.
Moreover, we can share the logits with the primary discriminative task and even exploit pretrained classifiers.

We can express the dense likelihood $p(\mathbf{x})$ by marginalizing out $\underline{\mathbf{y}}$:
\begin{equation}
    p(\mathbf{x}) = \sum_y p(\mathbf{y}, \mathbf{x}) = \frac{1}{Z} \, \sum_y \hat{p}(\mathbf{y}, \mathbf{x})   = \frac{1}{Z} \, \sum_i \exp{\mathbf{s}_i}.
\end{equation}
One could argue for detecting anomalies with $p(\mathbf{x})$ directly:
if a given input is unlikely under the $p(\mathbf{x})$, it should likely be an anomaly.
However, this approach may not work very well in practice due to tendency of maximum likelihood optimization towards over-generalization \cite{lucas19nips}.
In simple words, some outliers will have higher likelihood than the inliers \cite{serra20iclr,nalisnick19iclr}. 
We discourage such behaviour by minimizing the likelihood of negatives during the training \cite{hendrycks19iclr}.

Besides logit reinterpretation, we define the dataset posterior $P(d_{in}|\mathbf{x})$ as a non-linear transformation based on pre-logit activations $q_{\theta_1}(\mathbf{x})$ \cite{bevandic19gcpr}:
\begin{equation}
    P(d_{in}|\mathbf{x}) := \sigma(g_{\gamma}(q_{\theta_1}(\mathbf{x}))).
\end{equation}
In our case, the function $g$ is BN-ReLU-Conv1x1 of pre-logits, followed by a sigmoid non-linearity.
Anomalies can be detected solely with $P(d_{in}|\mathbf{x})$ \cite{devries18arxiv}: inlier samples should give rise to high posterior of the inlier dataset.
However, our experiments show that this is suboptimal compared to our hybrid approach.

\begin{figure}
    \centering
    \includegraphics[width=0.98\linewidth]{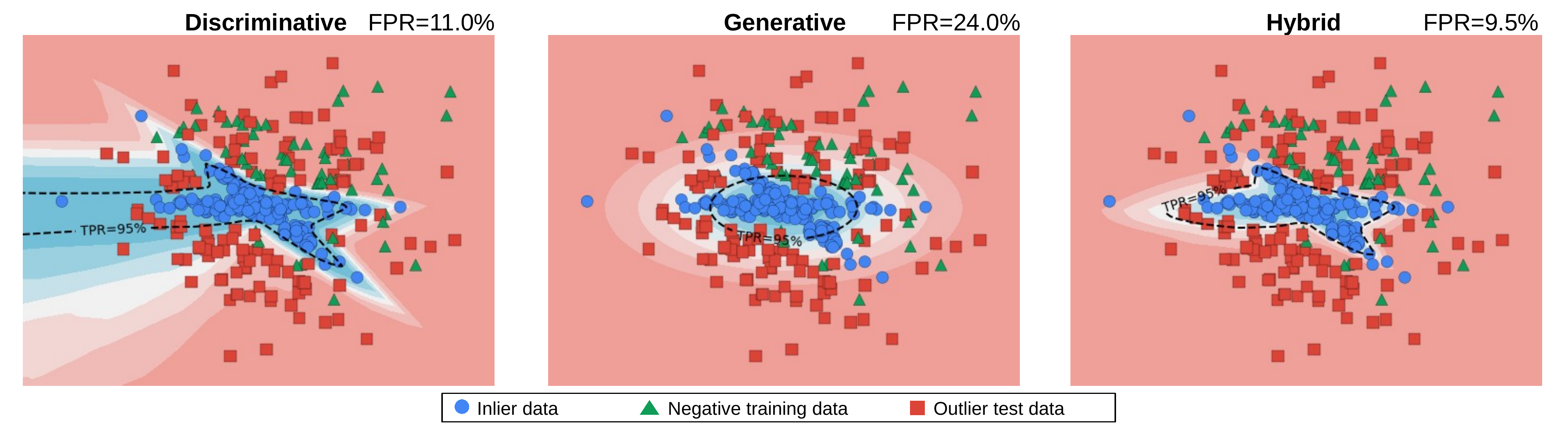}
    \caption{Anomaly detection on a toy dataset. The discriminative approach (left) models $P(d_{in}|\mathbf{x})$. It fails if the negative training dataset does not cover all modes of the test anomalies. The generative approach (middle) models $p(\mathbf{x})$. It often assigns high likelihoods to test anomalies due to over-generalization \cite{lucas19nips}. The hybrid approach achieves a synergy between discriminative and generative modelling}
    \label{fig:osr_poc}
\end{figure}

Fig. \ref{fig:osr_poc} illustrates shortcomings of generative and discriminative anomaly detectors on a toy problem.
Blue dots designate inlier data.
Green triangles designate the negative data used for training.
Red squares denote anomalous test data.
Discriminative detectors which model $P(d_{in}|\mathbf{x})$ can't differentiate inliers if the negative data seen during the training insufficiently covers the sample space (left).
On the other hand, generative detectors which model $p(\mathbf{x})$ tend to inaccurately distribute probability volume over sample space \cite{lucas19nips} (center).
Joining discriminative and generative approach into a hybrid detector we mitigate the aforementioned limitations (right).

We build our hybrid anomaly detector upon the discriminative dataset posterior $P(d_{in}|\mathbf{x})$ and the generative data likelihood $p(\mathbf{x})$.
We express a novel hybrid anomaly score as log-ratio 
between $P(d_{out}|\mathbf{x}) = 1 - P(d_{in}|\mathbf{x})$ and $p(\mathbf{x})$:
\begin{align}
\label{eq:sx}
    s(\mathbf{x}) & := \ln\frac{P(d_{out}|\mathbf{x})}{p(\mathbf{x})} = \ln P(d_{out}|\mathbf{x}) - \ln \hat{p}(\mathbf{x})  + \ln Z \\
    &\cong \ln P(d_{out}|\mathbf{x}) - \ln \hat{p}(\mathbf{x}).
\end{align}
We can neglect $Z$ since ranking performance \cite{hendrycks17iclr} is invariant to monotonic transformations such as taking a logarithm or adding a constant.
Other formulations of $s(\mathbf{x})$ may also be effective which is an interesting direction for future work.

\subsection{Dense Open-set Recognition based on Hybrid Anomaly Detection
\label{sec:ratio_train}}

\begin{figure}[ht]
    \centering
    \includegraphics[width=\linewidth]{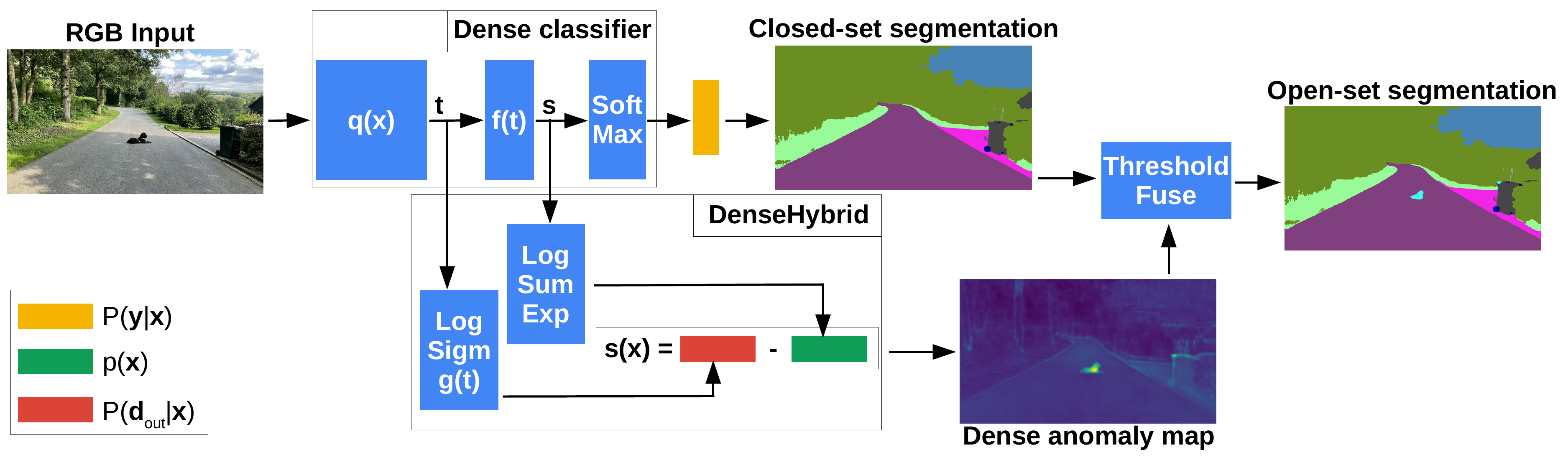}
    \caption{The proposed dense open-set recognition approach. Our anomaly score is a log-ratio of outputs derived from the hybrid model. We fuse the thresholded anomaly score with the closed-set segmentation map to obtain the open-set segmentation map}
    \label{fig:ratio}
\end{figure}

Figure \ref{fig:ratio} illustrates the inference with the proposed open-set recognition setup.
RGB input is fed to a hybrid dense model which produces pre-logit activations $\mathbf{t}$ and logits
$\mathbf{s}$.
Then, we obtain the closed-set class posterior $P(\mathbf{y}|\mathbf{x}) = \mathrm{softmax}(\mathbf{s})$ (designated in yellow) and the unnormalized data likelihood $\hat{p}(\mathbf{x})$ (designated in green).
A distinct head $g$ transforms pre-logits $\mathbf{t}$ 
into the dataset posterior $P(d_{out}|\mathbf{x})$.
The anomaly score $s(\mathbf{x})$ is a log-ratio between latter two distributions.
The resulting anomaly map is thresholded and fused with the discriminative output into the final dense open-set recognition map.

The developed hybrid model aims at achieving a synergy between generative and discriminative modelling.
However, the proposed hybrid interpretation requires 
specific training objectives.
Dense class posteriors require a discriminative loss over the inlier data $D_{in}$:
\begin{align}
\label{eq:cls}
    L_{\mathrm{cls}}(\theta) &= \mathbb{E}_{\mathbf{x}, \mathbf{y} \in D_{in}}[- \ln P(\mathbf{y}|\mathbf{x})] \\ &= -\, \mathbb{E}_{\mathbf{x}, \mathbf{y} \in D_{in}}[ \mathbf{s}_y] + \mathbb{E}_{\mathbf{x}, \mathbf{y} \in D_{in}}[ \ln \sum_i \exp{\mathbf{s}_i}].
\end{align}
The discriminative loss (\ref{eq:cls}) corresponds to the standard training in the closed world.
We introduce the negative data $D_{out}$ into the training procedure to ensure the desired behaviour of $P(d_{in}|\mathbf{x})$ and $p(\mathbf{x})$  \cite{hendrycks19iclr,bevandic19gcpr}.
Both distributions should yield low probability in negative pixels.
We propose to train $p(\mathbf{x})$ to maximize the likelihood in inliers and to minimize the likelihood in outliers.
We derive the upper bound of the desired loss as follows:
\begin{align}
L_{\mathbf{x}}(\theta) &= \mathbb{E}_{\mathbf{x} \in D_{in}}[- \ln p(\mathbf{x})] - \mathbb{E}_{\mathbf{x} \in D_{out}}[-\ln p(\mathbf{x})] \\
&= \mathbb{E}_{\mathbf{x} \in D_{in}}[- \ln \hat{p}(\mathbf{x})] + \cancel{\ln{Z}} - \mathbb{E}_{\mathbf{x} \in D_{out}}[-\ln \hat{p}(\mathbf{x})] - \cancel{\ln{Z}}  \\
&= - \, \mathbb{E}_{\mathbf{x} \in D_{in}}\left[\ln\sum_i\exp(\mathbf{s}_i)\right]  + \, \mathbb{E}_{\mathbf{x} \in D_{out}}\left[\ln\sum_i\exp(\mathbf{s}_i)\right] \\
&\leq - \, \mathbb{E}_{\mathbf{x}, \mathbf{y} \in D_{in}}[\mathbf{s}_y]  + \, \mathbb{E}_{\mathbf{x} \in D_{out}}[\ln\sum_i\exp(\mathbf{s}_i)].
\label{eq:x_d}
\end{align}
Note that we eschew the backpropagation into the normalization constant $Z$, and derive the upper bound according to the following inequality:
\begin{equation}
\label{eq:ineq}
\ln\sum_i\exp{\mathbf{s}_i} \geq \max_i \mathbf{s}_i \geq \mathbf{s}_y.
\end{equation}
Proof of inequality (\ref{eq:ineq}) can be easily derived by recalling that log-sum-exp is a smooth upper bound of the max function.
By comparing the standard classification loss (\ref{eq:cls}) and the upper bound (\ref{eq:x_d}) we realize that minimizing the standard classification loss increases $p(\mathbf{x})$ for inlier pixels.
Indeed, minimizing the negative logarithm of softmax output increases the value of logit for the correct class.

Alternatively, $p(\mathbf{x})$ could be trained only on inliers \cite{salakhutdinov09aistats,du19neurips,grathwohl20iclr}.
This would require sample hallucination via MCMC sampling and back-propagation into the corresponding approximation of $Z$.
Such procedure is infeasible for large images.
Consequently, we choose to deal with negative samples instead of hallucinated ones and optimize
the proposed loss $L_{\mathbf{x}}(\theta)$.

We train the dataset posterior $P(d_{in}|\mathbf{x})$ with the standard discriminative loss \cite{bevandic19gcpr}:
\begin{equation}
\label{eq:d_x}
    L_{\mathbf{d}}(\theta, \gamma) = \mathbb{E}_{\mathbf{x} \in D_{in}}[ - \ln P(d_{in}|\mathbf{x})] + \mathbb{E}_{\mathbf{x} \in D_{out}}[ - \ln(P(d_{out}|\mathbf{x}))].
\end{equation}
By joining losses $L_{\mathrm{cls}}$, $L_{\mathbf{x}}$ and $L_{\mathbf{d}}$ we obtain the final loss:
\begin{multline}
\label{eq:final_loss}
    L(\theta, \gamma) = - \mathbb{E}_{\mathbf{x},\mathbf{y} \in D_{in}}[\ln P(\mathbf{y}|\mathbf{x}) + \ln P(d_{in}|\mathbf{x})] \\- \beta \cdot \mathbb{E}_{\mathbf{x} \in D_{out}}[ \ln( P(d_{out}|\mathbf{x})) - \ln \hat{p}(\mathbf{x})].
\end{multline}
Hyperparameter $\beta$ controls the impact of negative data to the primary classification task.
Note that the final loss (\ref{eq:final_loss}) omits the first term from $L_{\mathbf{x}}$ (\ref{eq:x_d}) in positive pixels.
We choose to do so since $\hat{p}(\mathbf{x})$ is implicitly optimized through $L_{\mathrm{cls}}$.

\begin{figure}
    \centering
    \includegraphics[width=\linewidth]{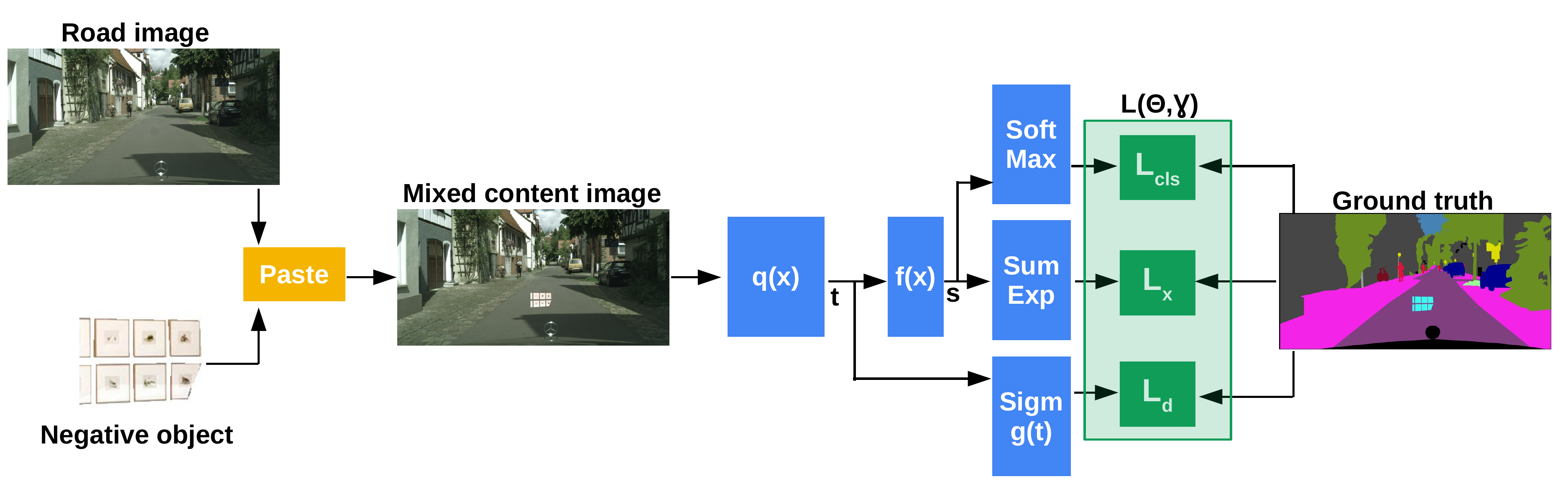}
    \caption{
    The training procedure of the proposed open-set recognition model.
    Mixed-content images are fed to the open-set model with three outputs. Each output is optimized according to the compound loss (\ref{eq:final_loss})
    }
    \label{fig:ratio_train}
\end{figure}

Figure \ref{fig:ratio_train} illustrates the described training procedure of the proposed open-set recognition model.
We prepare the training images by pasting the negative instances atop the standard training images.
The resulting mixed-content image \cite{bevandic19gcpr} is fed to the hybrid model.
We obtain the classification output $P(\mathbf{y}|\mathbf{x})$ with softmax.
The unnormalized likelihood $\hat{p}(\mathbf{x})$ is obtained through sum-exp operator.
We recover $p(d_{in}|\mathbf{x})$ by branching from pre-logit activations. 
The model outputs are trained by applying the dicriminative loss $L_\mathrm{cls}$  (\ref{eq:cls}), likelihood loss $L_\mathbf{x}$ (\ref{eq:x_d}) and dataset posterior loss $L_\mathbf{d}$ (\ref{eq:d_x}).
As proposed, these losses are conveniently joined into a single loss $L(\theta, \gamma)$ (\ref{eq:final_loss}).

\section{Measuring Dense Open-set Performance}
\label{sec:openiou}
Test datasets for anomaly segmentation either exclusively measure the performance of anomaly detectors \cite{pinggera16iros,chan21arxiv} or simply report the classification performance \cite{blum21ijcv}.
In the latter case, the reported drop in segmentation performance is usually negligible and is explained away by allocation of model capacity for the anomaly detection.
We will show that the real impact of anomaly detector on the segmentation performance can be clearly seen only in the open world.
Also, the impact is more severe than the small performance drop visible in the closed world.

To properly measure open-set recognition performance, we first select threshold at which the anomaly detector achieves TPR of 95\%.
This ensures high safety standards for the recognition model.
Then, we override the classification in pixels which raise concern according to the thresholded anomaly map.
The resulting recognition map has $K+1$ labels.
We compute the recognition performance in open-world using open intersection over union (open-IoU).
For the $k$-th class we can compute the proposed open-IoU as:
\begin{equation}
    \text{open-IoU}_k = \frac{\mathrm{TP}_k}{\mathrm{TP}_k + \mathrm{FP}^\text{ow}_k + \mathrm{FN}^\text{ow}_k}, \, \mathrm{FP}^\text{ow}_{k} = \sum_{\overset{i\neq k}{i=1}}^{K+1} \mathrm{FP}_k^i, \, \mathrm{FN}^\text{ow}_{k} = \sum_{\overset{i\neq k}{i=1}}^{K+1} \mathrm{FN}_k^i 
\end{equation}
Different that the standard IoU formulation, open-IoU also takes into account false positives and false negatives caused by imperfect anomaly detector.
However, we still average open-IoU over $K$ inlier classes.
This means that a recognition model which uses a perfect anomaly detector would match segmentation performance in the closed world.
This property would not be preserved if we averaged IoU over K+1 classes.

\begin{figure}
    \centering
    \includegraphics[width=0.9\linewidth]{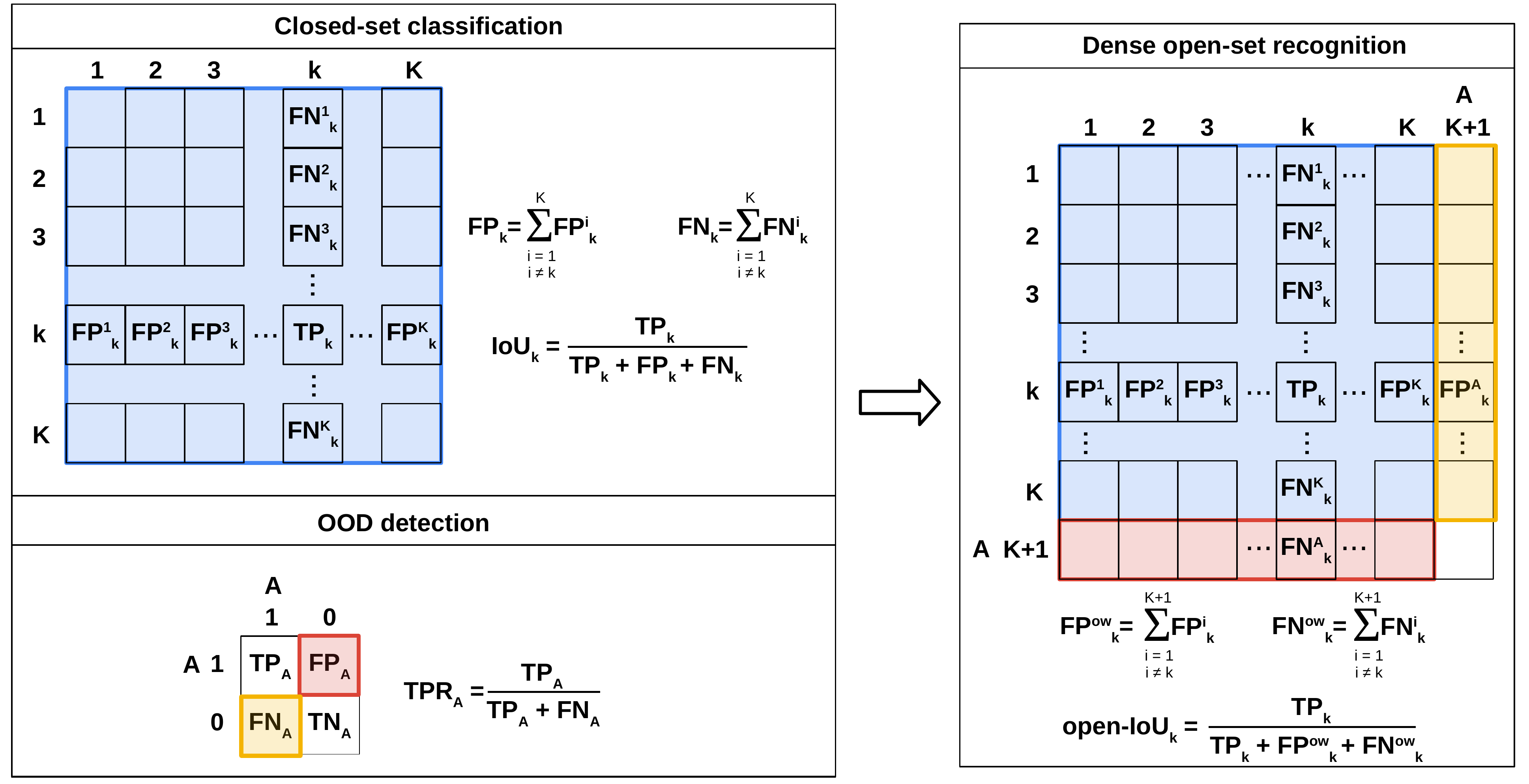}
    \caption{
    The proposed open intersection over union (open-IoU) takes into account missclassifications in anomalous pixels to accurately measure dense recognition performance in open world}
    \label{fig:conf_mat}
\end{figure}

Figure \ref{fig:conf_mat} (right) shows the open world confusion matrix.
Imperfect anomaly detection impacts recognition performance through increased false positives (designated in yellow) and false negatives (designated in red).
Difference between closed mIoU and averaged open-IoU over $K$ inlier classes reveals the performance hit due to inaccurate anomaly detection.

Measuring performance using the proposed open-IoU requires datasets with K+1 labels.
Creating such taxonomy requires substantial resources.
Currently, only StreetHazards \cite{hendrycks19arxiv} offers appropriate taxonomy for measuring open-IoU.

\section{Experiments}

We report dense anomaly detection and open-set recognition performance of the proposed DenseHybrid approach, and compare them with the state of the art.
We also explore influence of distance, show computational requirements of the proposed module, and ablate the design choices.

\subsection{Benchmarks and Datasets}

We evaluate performance on standard benchmarks for dense anomaly detection.
Fishyscapes \cite{blum21ijcv} considers urban scenarios on a subset of LostAndFound \cite{pinggera16iros} and on Cityscapes validation images with pasted anomalies (FS Static).
SegmentMeIfYouCan (SMIYC) \cite{chan21arxiv} moves away from anomaly injection.
Instead, appropriate images are collected from the real world and grouped based on the anomaly size into AnomalyTrack (large) and ObstacleTrack (small).
Additionally, the benchmark encapsulates all LostAndFound images.
Unfortunately, both benchmarks only have binary labels which makes them insufficient for measuring the recognition performance as proposed in Sec.\ \ref{sec:openiou}.
StreetHazards \cite{hendrycks19arxiv} is a synthetic dataset created by CARLA virtual environment.
The simulated environment enables smooth anomaly injection and low-cost label extraction.
Consequently, the dataset contains $K+1$ labels which makes it suitable for measuring both anomaly detection and dense recognition.

\subsection{Dense Anomaly Detection}
\label{sec:res_anomdet}
Table \ref{table:smiyc} shows performance of 
the proposed hybrid 
anomaly detector on the SMIYC benchmark \cite{chan21arxiv}.
DenseHybrid outperforms contemporary approaches on both AnomalyTrack and ObstacleTrack by a wide margin.
Also, the proposed anomaly detector 
achieves the best FPR on LostAndFound.

\begin{table}[h]
\begin{center}
\caption{Performance evaluation on the SMIYC benchmark \cite{chan21arxiv}.
DenseHybrid outperforms contemporary approaches on Anomaly and Obstacle track by a wide margin, while also achieving the best FPR on LostAndFound
}
\label{table:smiyc}
\begin{tabular}{lcccccccc}
\hline\hline
\multicolumn{1}{l}{\multirow{2}{*}{Method}} & \multicolumn{1}{c}{\multirow{2}{*}{Aux}} & \multicolumn{1}{c|}{\multirow{2}{*}{Img}} & \multicolumn{2}{c|}{AnomalyTrack} & \multicolumn{2}{c|}{ObstacleTrack} & \multicolumn{2}{c}{LAF-noKnown}\\
\multicolumn{1}{l}{} & \multicolumn{1}{c}{data} & \multicolumn{1}{c|}{rsyn.} & AP & \multicolumn{1}{c|}{$\mathrm{FPR}_{95}$} & AP & \multicolumn{1}{c|}{$\mathrm{FPR}_{95}$}  & AP & \multicolumn{1}{c}{$\mathrm{FPR}_{95}$}  \\\hline \hline
SynBoost \cite{biase21cvpr} & \cmark& \cmark & 56.4 & 61.9 &  71.3 & 3.2 &  81.7& 4.6\\
Image Resyn. \cite{lis19iccv} & \xmark& \cmark& 52.3 & 25.9 & 37.7 & 4.7 & 57.1& 8.8\\
JSRNet \cite{vojir21iccv} & \xmark&  \cmark & 33.6& 43.9  & 28.1 & 28.9 & 74.2 & 6.6\\
Road Inpaint. \cite{lis20arxiv} & \xmark& \cmark & - & - & 54.1 & 47.1 & \textbf{82.9} & 35.8\\
Embed.\ Dens.\ \cite{blum21ijcv} & \xmark&  \xmark & 37.5 & 70.8  & 0.8 & 46.4 & 61.7 &10.4 \\
ODIN \cite{liang18iclr} & \xmark&  \xmark & 33.1 & 71.7 & 22.1 & 15.3 & 52.9 &  30.0 \\
MC Dropout \cite{kendall17nips} & \xmark& \xmark & 28.9 & 69.5 & 4.9 & 50.3 & 36.8 & 35.6 \\
Max softmax \cite{hendrycks17iclr} & \xmark&  \xmark & 28.0 & 72.1 & 15.7 & 16.6  & 30.1 & 33.2 \\
Mahalanobis \cite{lee18nips} & \xmark&  \xmark & 20.0 & 87.0  &  20.9 & 13.1 & 55.0 & 12.9\\
Void Classifier \cite{blum21ijcv} & \cmark& \xmark & 36.6 & 63.5 & 10.4 & 41.5 & 4.8  & 47.0 \\
\hline
DenseHybrid (ours) & \cmark&  \xmark & \textbf{78.0} & \textbf{9.8}  & \textbf{87.1} & \textbf{0.2} & 78.7 & \textbf{2.1} \\ \hline
\end{tabular}
\end{center}
\end{table}

Table \ref{table:fishy} shows performance of the proposed DenseHybrid on Fishyscapes \cite{blum21ijcv}.
Our anomaly detector achieves the best results on FS LostAndFound, and the best FPR on FS Static.
We achieve these results while having negligible impact on classification task in closed-world.
However, in the next section we show that the impact of anomaly detection to recognition performance is much more significant than in the closed world.
\begin{table}[h]
\begin{center}
\caption{
Performance evaluation on the Fishyscapes benchmark \cite{blum21ijcv}.
DenseHybrid achieves the best performance on FS LostAndFound and the best FPR on FS Static}
\label{table:fishy}
\begin{tabular}{lccccccc}
\hline\hline
\multicolumn{1}{l}{\multirow{2}{*}{Method}} & \multicolumn{1}{c}{\multirow{2}{*}{Aux}} & \multicolumn{1}{c|}{\multirow{2}{*}{Img}} & \multicolumn{2}{c|}{LostAndFound} & \multicolumn{2}{c|}{Static} & Closed world\\
\multicolumn{1}{l}{} & \multicolumn{1}{c}{data} & \multicolumn{1}{c|}{rsyn.} & AP & \multicolumn{1}{c|}{$\mathrm{FPR}_{95}$} & AP & \multicolumn{1}{c|}{$\mathrm{FPR}_{95}$} & Cityscapes mIoU \\\hline \hline
SynBoost \cite{biase21cvpr} & \cmark& \cmark & 43.2 & 15.8 &  72.6 & 18.8 & \textbf{81.4}\\
Image Resyn. \cite{lis19iccv} & \xmark& \cmark& 5.7 & 48.1 & 29.6 & 27.1 & 81.4\\
Standardized ML \cite{jung21iccv} & \xmark&  \xmark & 31.1 & 21.5 & 53.1 & 19.6 & 80.3\\
Embed.\ Dens.\ \cite{blum21ijcv} & \xmark&  \xmark & 4.7 & 24.4  & 62.1 & 17.4  & 80.3\\
Max softmax \cite{hendrycks17iclr} & \xmark&  \xmark & 1.77 & 44.9 & 12.9 & 39.8  & 80.3\\
Dirichlet prior \cite{malinin18nips} & \cmark& \xmark & 34.3 & 47.4 & \textbf{84.6} & 30.0 & 70.5\\ 
OOD Head \cite{bevandic19gcpr} & \cmark& \xmark & 30.9 & 22.2 & 84.0 & 10.3 & 77.3\\ 
Void Classifier \cite{blum21ijcv} & \cmark& \xmark & 10.3 & 22.1 & 45.0 & 19.4 & 70.4\\
Mutual information \cite{mukhoti18arxiv} & \cmark& \xmark & 9.8 & 38.5 & 48.7 & 15.5 & 73.8\\\hline
DenseHybrid (ours) & \cmark&  \xmark & \textbf{43.9} & \textbf{6.2} & 72.3 & \textbf{5.5} & \textbf{81.0} \\  \hline
\end{tabular}
\end{center}
\end{table}

Table \ref{tbl:distance} explores sensitivity of anomaly detection with respect to distance from the camera.
We perform all these experiments on LostAndFound since it includes disparity maps.
Still, due to errors in available disparities, we limit our analysis to the first 50 meters from the camera.
The proposed DenseHybrid approach achieves accurate results even at large distances from the vehicle.

\begin{table}[h]
\begin{center}
\caption{Anomaly detection performance at different distances from camera.
Our DenseHybrid based on DeeplabV3+ with WRN38 backbone \cite{zhu19cvpr} accurately detects anomalies at different ranges}
\label{tbl:distance}
\begin{tabular}{lcccccccccc}
\hline \hline
 \multirow{2}{*}{Method}& \multirow{2}{*}{Metric}  & \multicolumn{9}{|c}{Range in meters} \\
 &  & \multicolumn{1}{|c}{5-10} & 10-15 & 15-20 & 20-25 & 25-30 & 30-35 & 35-40 & 40-45 & 45-50 \\ \hline \hline
 \multirow{2}{*}{Max-softmax \cite{hendrycks17iclr}} & AP & 28.7 & 28.8 & 26.0 & 25.1 & 29.0 & 26.2 & 29.6 & 31.7 & 33.7\\
& $\mathrm{FPR}_{95}$  & 16.4 & 29.7 & 28.8 &44.2 & 41.3 & 47.8 & 44.7 & 43.2 & 45.3 \\ \hline
 \multirow{2}{*}{Max-logit \cite{hendrycks19arxiv}} & AP & 76.1 & 73.9 & 78.2 & 69.6 & 72.6 & 70.2 & 71.0 & 74.0 & 73.9\\
& $\mathrm{FPR}_{95}$  & 5.4 & 16.2 & 5.9 & 12.8 & 9.5 & 10.0 & 9.8 & 9.8 & 11.0 \\ \hline
\multirow{2}{*}{SynBoost \cite{biase21cvpr}} & AP & \textbf{93.7} & 78.7 & 76.9 & 70.0 & 65.6 & 58.5 & 59.8 & 60.0 & 53.3\\
 & $\mathrm{FPR}_{95}$ & \textbf{0.2} & 17.7 & 25.0 & 23.3 &18.8  & 27.4  & 25.4 &25.8  & 29.9 \\ \hline
 \multirow{2}{*}{DenseHybrid (ours)}& AP & 90.7 & \textbf{89.8} & \textbf{92.9} & \textbf{89.1} & \textbf{89.5} & \textbf{87.7} & \textbf{85.0} & \textbf{85.6} & \textbf{82.1} \\
& $\mathrm{FPR}_{95}$ & 0.3 & \textbf{1.1} & \textbf{0.6} & \textbf{1.4} & \textbf{1.4} & \textbf{2.5} & \textbf{3.7} & \textbf{4.7} & \textbf{6.3} \\\hline
\end{tabular}
\end{center}
\end{table}

\subsection{Dense Open-set Recognition}
\label{sec:res_rec}
By fusing a properly thresholded anomaly detector with the dense classifier, we obtain a dense open-set recognition model (Fig.\ \ref{fig:ratio}).
The resulting model detects anomalous scene parts, while correctly classifying the rest of the scene.

To measure the dense recognition performance, we create two test folds based on towns t5 and t6 from StreetHazards test. 
Then, we select anomaly threshold on t6 and use it to measure the proposed open-mIoU on t5.
We switch the folds and repeat the procedure.
We compute the weighted average based on image count to obtain the final test set open-mIoU.

Table \ref{table:osr_sh} shows performance of our dense recognition models on StreetHazards.
The left part of the table considers anomaly detection where DenseHybrid achieves the best performance.
The right part of the table considers dense recognition performance.
Our model outperforms other contemporary approaches despite lower classification performance in the closed world.
Note that the performance drop between the closed and the open set is significant.
The models achieve over 60\% mIoU in closed world while the open world performance peeks at 46\%.
Hence, we conclude that even the best anomaly detectors are still insufficient for matching the closed world performance in open-world.
Researchers should strive to close this gap in order to improve the safety of recognition systems in the real world. 

\begin{table}[h]
\begin{center}
\caption{Performance evaluation on StreetHazards \cite{hendrycks19arxiv}.
DenseHybrid achieves the best anomaly detection performance.
The corresponding open-set recognition model yields the best performance measured by open-mIoU (Sec.\ \ref{sec:openiou})}
\label{table:osr_sh}
\begin{tabular}{lcccccccc}
\hline \hline
\multirow{2}{*}{Method} & \multicolumn{1}{c|}{Aux.} & \multicolumn{3}{c|}{Anomaly detection} & \multicolumn{1}{c|}{Closed world} & \multicolumn{3}{c}{Open world}\\
  & \multicolumn{1}{c|}{data}  &  AP         & $\mathrm{FPR}_{95}$       & \multicolumn{1}{c|}{AUC}     & \multicolumn{1}{c|}{$\overline{\mathrm{IoU}}$} &  o-$\overline{\mathrm{IoU}}$-t5 & o-$\overline{\mathrm{IoU}}$-t6 & o-$\overline{\mathrm{IoU}}$\\ \hline \hline
SynthCP \cite{xia20eccv} & \xmark &  9.3           & 28.4          & 88.5     & - &  -  & - & -\\
Dropout \cite{kendall17nips}\cite{xia20eccv} &  \xmark &  7.5           & 79.4          & 69.9   & - &  - & - & -\\
TRADI \cite{franchi20eccv} & \xmark &  7.2           & 25.3          & 89.2        & - & -  & - & -\\
OVNNI \cite{franchi20arxiv} & \xmark & 12.6  & 22.2 & 91.2  & 54.6 & - & - & -\\
SO+H \cite{grcic21visapp}& \xmark & 12.7  & 25.2 & 91.7 & 59.7 & - & - & -\\
DML \cite{cen21iccv} & \xmark  & 14.7  & 17.3  &  93.7 &  - & - & - & -\\
MSP \cite{hendrycks17iclr} & \xmark  &  7.5   &   27.9  & 90.1 &  65.0   &  32.7 & 40.2 & 35.1\\
ML \cite{hendrycks19arxiv} & \xmark  & 11.6   &  22.5    & 92.4  & 65.0   &  39.6 & 44.5 & 41.2\\
ODIN \cite{liang18iclr}&   \xmark   &    7.0       & 28.7    &   90.0  & 65.0 & 26.4 & 33.9 & 28.8\\
ReAct \cite{sun21neurips} & \xmark & 10.9  & 21.2 & 92.3 &  62.7 & 33.0 & 36.2  & 34.0\\
Energy \cite{liu20neurips}& \cmark &  12.9  & 18.2  & 93.0 & 63.3 & 41.7 & 44.9 & 42.7\\
Outlier Exposure \cite{hendrycks19iclr} & \cmark  &  14.6   &  17.7  & 94.0  & 61.7   &  43.7 & 44.1 & 43.8\\
OOD-Head \cite{bevandic19gcpr}  & \cmark   &  19.7 &  56.2   & 88.8 &  \textbf{66.6} & 33.7 & 34.3 & 33.9\\
OH*MSP \cite{bevandic21arxiv} & \cmark  & 18.8  & 30.9 & 89.7 &   \textbf{66.6}  & 43.3 & 44.2 & 43.6\\\hline
DenseHybrid (ours) & \cmark &  \textbf{30.2}  & \textbf{13.0} & \textbf{95.6}  & 63.0 & \textbf{46.1} & \textbf{45.3} & \textbf{45.8} \\\hline
\end{tabular}
\end{center}
\end{table}

\begin{figure}
    \centering
    \includegraphics[width=\linewidth]{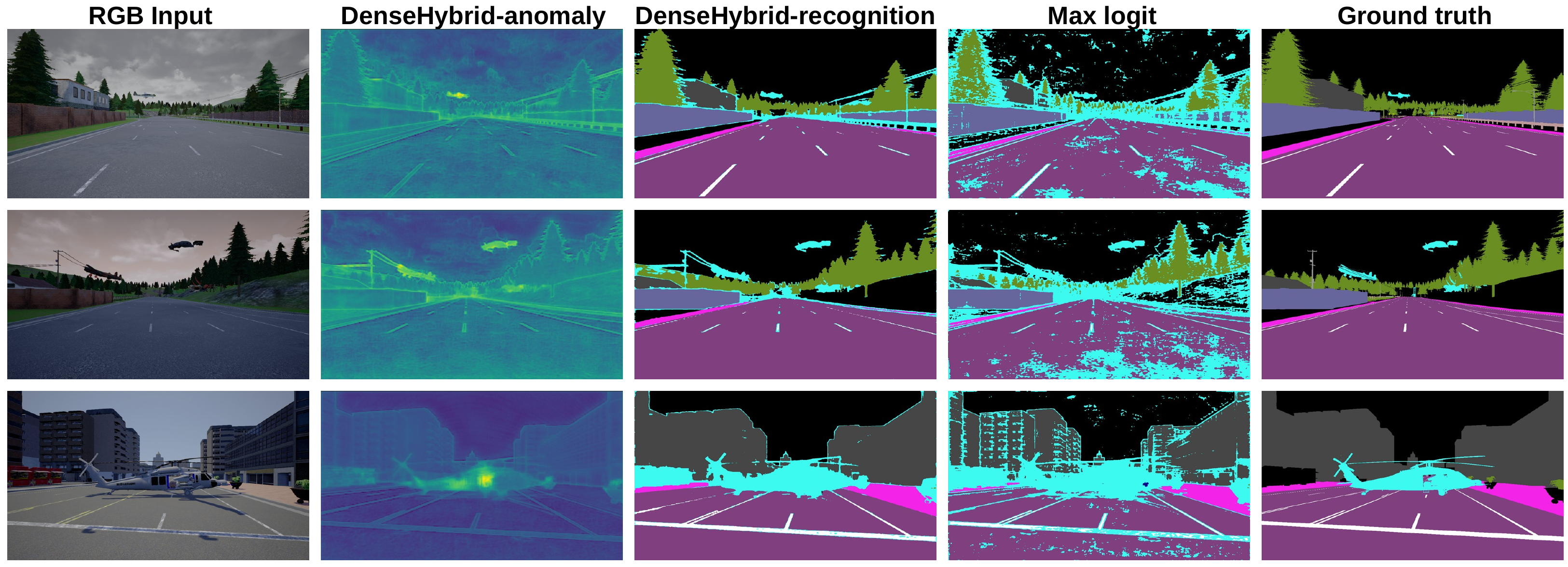}
    \caption{Visualisation of dense open-set recognition performance on StreetHazards.
    DenseHybrid significantly outperforms the max-logit baseline \cite{hendrycks19arxiv}}
    \label{fig:sh_results}
\end{figure}

Figure \ref{fig:sh_results} visualises dense anomaly and recognition maps on StreetHazards.
Our recognition model significantly outperforms the max-logit baseline \cite{hendrycks19arxiv}.

\subsection{Inference speed}
Table \ref{tbl:speed} shows computational overhead of the proposed DenseHybrid anomaly detector over the baseline segmentation model on two megapixels images.
DenseHybrid has negligible computational overhead of 0.1 GFLOPs and 2.8ms.
Our results are averaged 
over 200 runs on NVIDIA RTX3090.
These experiments also suggest that 
image resynthesis is not applicable 
for real-time inference.
\begin{table}[h]
\begin{center}
\caption{Computational overhead of the proposed DenseHybrid anomaly detector when inferring with RTX3090 on two megapixel images
}
\label{tbl:speed}
\begin{tabular}{lcccc}
\hline \hline
Method & \multicolumn{1}{c|}{Resynth.} & \multicolumn{1}{c|}{Infer. time (ms)} & \multicolumn{1}{c|}{Frames per sec.} & GFLOPs \\\hline \hline
SynBoost \cite{biase21cvpr} & \cmark & 1055.5 & $<$1 & - \\
SynthCP \cite{xia20eccv} & \cmark & 146.9 & $<$1 & 4551.1 \\
LDN-121 \cite{kreso21tits} & \xmark & 60.9 & 16.4 & 202.3 \\ 
LDN-121 + SML \cite{jung21iccv} & \xmark & 75.4 & 13.3 & 202.6\\
LDN-121 + DenseHybrid (ours) & \xmark & \textbf{63.7}  & \textbf{15.7} & \textbf{202.4} \\\hline
\end{tabular}
\end{center}
\end{table}

\subsection{Impact of anomaly detector design}
Table \ref{table:abl_sx} compares the proposed DenseHybrid approach with its generative and discriminative components -- $\hat{p}(\mathbf{x})$ and $P(d_{in}|\mathbf{x})$.
The hybrid anomaly score based on the ratio of these two distributions outperforms each of the two components.
The results are averaged over the last three epochs.
\begin{table}[h]
\begin{center}
\caption{Validation of DenseHybrid components on Fishyscapes validation set}
\label{table:abl_sx}
\begin{tabular}{lcccc}
\hline \hline
\multirow{2}{*}{Anomaly detector} & \multicolumn{2}{c|}{FS LostAndFound} & \multicolumn{2}{c}{FS Static} \\
 & AP & \multicolumn{1}{c|}{$\mathrm{FPR}_{95}$}  & AP & $\mathrm{FPR}_{95}$ \\ \hline \hline
Discriminative $(1-P(d_{in}|\mathbf{x}))$ & 42.9 $\pm$ 4.2 & 42.1 $\pm$ 7.0 & 47.8 $\pm$ 5.0 & 41.6 $\pm$ 8.3 \\
Generative $\hat{p}(\mathbf{x})$ & 60.5 $\pm$ 2.6&  7.4 $\pm$ 0.8 & 54.2 $\pm$ 2.1 & 6.2 $\pm$ 0.7\\
Hybrid $(1-P(d_{in}|\mathbf{x}))/\hat{p}(\mathbf{x})$  & \textbf{63.8} $\pm$ 2.9 & \textbf{6.1} $\pm$ 0.7 & \textbf{60.0} $\pm$ 2.0 & \textbf{4.9} $\pm$ 0.6 \\ \hline
\end{tabular}
\end{center}
\end{table}

\subsection{Implementation details}
We adapt the standard segmentation networks \cite{kreso21tits,zhu19cvpr} to enable co-operation with our hybrid anomaly detector.
We append an additional branch $g_\gamma$ which is in our case BN-ReLU-Conv1x1. The additional branch computes the discriminative anomaly output.
We obtain generative anomaly output by computing sum of exponentiated logits.
We build our recognition models based on dense classifiers.
We fine-tune all our models on mixed content images with pasted negative instances from ADE20k.
In the case of SMIYC we fine-tune LDN-121 \cite{kreso21tits} for 10 epochs on images from Cityscapes \cite{cordts16cvpr}, Vistas \cite{neuhold17iccv} and Wilddash2 \cite{zendel18eccv}.
In the case of Fishyscapes we use DeepLabV3+ with WideResNet38 \cite{zhu19cvpr}.
We fine-tune the model for 10 epochs on Cityscapes.
We train LDN-121 on StreetHazards for 120 epochs in closed world and then fine-tune the recognition model on mixed-content images.
Other details are available in the supplement.

\section{Conclusion}

Discriminative and generative approaches to dense anomaly detection assume different failure modes.
We propose to achieve a synergy of these two approaches by fusing the data posterior and the data likelihood derived from the standard discriminative model.
The proposed hybrid setup relies on unnormalized distributions.
Hence, we try to eschew evaluation of the intractable normalization constant both during  training and inference.
The proposed DenseHybrid architecture yields state-of-the-art performance on the standard anomaly segmentation benchmarks as well as competitive dense recognition performance in the open world.
The latter is measured with the novel open-mIoU score which takes into account classification in both inliers and anomalous pixels.
Future work should focus on reducing the revealed performance gap between closed-world and open-world recognition in order to improve the progress toward safe autonomous driving systems.

\section*{Acknowledgements}
This work has been supported by Croatian Science Foundation grant IP-2020-02-5851 ADEPT, as well as by European Regional Development Fund grants KK.01.2.1.02.0119 DATACROSS and KK.01.2.1.02.0119 A-Unit co-funded by Gideon Brothers ltd. We thank Marin Oršić for insightful discussions.

%
%
\bibliographystyle{splncs04}
\bibliography{main}

\newpage
\section{Supplement}

\subsection{Limitations}
We term our method hybrid since it optimizes two different training objectives:
i) dense discrimination between inliers and negatives, and
ii) high likelihood of inliers and low likelihood of negative data.
 It may seem that our method can generate samples
due to likelihood evaluation being
a standard feature of generative models (except GANs).
However, our formulation is not suitable for sample generation due to dealing with unnormalized distributions.
This would require MCMC sampling which can not be performed at large resolutions, at least not with known techniques.
Even if sample generation was feasible, the resulting approach would likely be too slow for real-time inference as shown by other image resynthesis approaches \cite{lis19iccv,biase21cvpr}.

\subsection{Impact of Known Inliers on Anomaly Detection}
Many real-world deployments of autonomous systems work in environment with limited variety (e.g. warehouses or industrial plants).
Such environments usually have perfectly aligned training and test distributions.
Still, anomalous objects can occur.
We show that anomaly detection performance in such cases is significantly more easier.
Table \ref{table:cityval} shows such setup.
We used two DLv3+ segmentation networks.
The first one is trained on Cityscapes train while the second one is trained on Cityscapes train and val splits.
Since the Fishyscapes Static dataset is created by pasting negative objects atop Cityscapes val images, we can measure anomaly detection performance when the inlier instances from test set are known.
We see that the average precision of anomaly detection is drastically improved from 60\% to 89\%.
This indicates that the DenseHybrid anomaly detector is feasible for scenarios with limited scene variety.
\begin{table}[h]
\begin{center}
\caption{
Impact of known inlier instances on anomaly detection performance.
Results indicate that DenseHybrid is feasible for scenarios with limited scene variety}
\label{table:cityval}
\begin{tabular}{lccc}
\hline \hline
\multirow{2}{*}{Training splits} & \multicolumn{2}{c|}{FS Static} & Closed world \\
 &  AP & \multicolumn{1}{c|}{$\mathrm{FPR}_{95}$} & Cityscapes val mIoU\\ \hline \hline
train  & 60.0$\pm$2.0 & 4.9$\pm$0.6 &  81.0\\
train+val &  88.5$\pm$0.8 & 1.1$\pm$0.1 & 89.9 \\ \hline
\end{tabular}
\end{center}
\end{table}

\subsection{More results}

Table \ref{table:osr_more} shows the dense open-set recognition performance with generative and discriminative anomaly detectors on the StreetHazards dataset.
The proposed hybrid anomaly score based on the ratio of these two distributions outperforms each of the two components.
Note that generative and discriminative outputs are trained using the same mixed-content images.
\begin{table}[h]
\begin{center}
\caption{Performance evaluation on StreetHazards.
DenseHybrid anomaly score based on the ratio of generative and discriminative distributions outperforms each of the two components}
\label{table:osr_more}
\begin{tabular}{lccccccccc}
\hline \hline
\multirow{2}{*}{Method} & \multicolumn{3}{c|}{Anomaly detection} & \multicolumn{3}{c|}{Closed world} & \multicolumn{3}{c}{Open world}\\
  &  AP         & $\mathrm{FPR}_{95}$       & \multicolumn{1}{c|}{AUC}    &  $\overline{\mathrm{IoU}}$-t5& \multicolumn{1}{c}{$\overline{\mathrm{IoU}}$-t6} & \multicolumn{1}{c|}{$\overline{\mathrm{IoU}}$} &  o-$\overline{\mathrm{IoU}}$-t5 & o-$\overline{\mathrm{IoU}}$-t6 & o-$\overline{\mathrm{IoU}}$\\ \hline \hline
Generative &  30.0  & 13.3 & 95.5  & 65.6  &  61.6  & 63.0 & 45.6 & 45.2 & 45.5 \\
Discriminative &  23.3  & 20.5 & 93.1  & 65.6  &  61.6  & 63.0 & 36.9 & 35.2 & 36.4 \\
DenseHybrid (ours) &  \textbf{30.2}  & \textbf{13.0} & \textbf{95.6}  & 65.6  &  61.6  & 63.0 & \textbf{46.1} & \textbf{45.3} & \textbf{45.8} \\\hline
\end{tabular}
\end{center}
\end{table}

\subsection{Implementation Details}

We create mixed content training samples by pasting negative data atop the inlier crops.
The inlier crops are obtained by jittering images in range [0.5, 2], applying random horizontal flip, and taking random square crop of size 768.
We use ADE20k instances as negative content.
We resize the negative image to 768 pixels, take a random jittered crop of size 384.
Then, we paste two instances from the negative crop at random position atop the inlier crop.
For LDN-121, we use batch size 16.
We use Adam optimizer with the initial learning rate $10^{-5}$.
The learning rate is decayed through a cosine schedule down to $10^{-7}$.
We set the loss modulation hyper-parameter $\beta$ to 0.03.
For DLv3+ we use batch size 8 and inlier crops of 512.
We use Adam optimizer with learning rate $10^{-6}$ and do not decay it.
We set the loss modulation hyper-parameter $\beta$ to 0.01.

In the case of Fishyscapes, we fine-tune DLv3+ with a WRN38 backbone pre-trained by NVIDIA \cite{zhu19cvpr}.
However, due to hardware limitations we could not train it from scratch to achieve the desired robustness required for SMIYC.
Hence, we opted for LDN-121 as an efficient alternative which can be trained on a single GPU.
Using the bigger DLv3+ would additionally improve results on SMIYC.


\subsection{Visualizations}

\begin{figure}
    \centering
    \includegraphics[width=\linewidth]{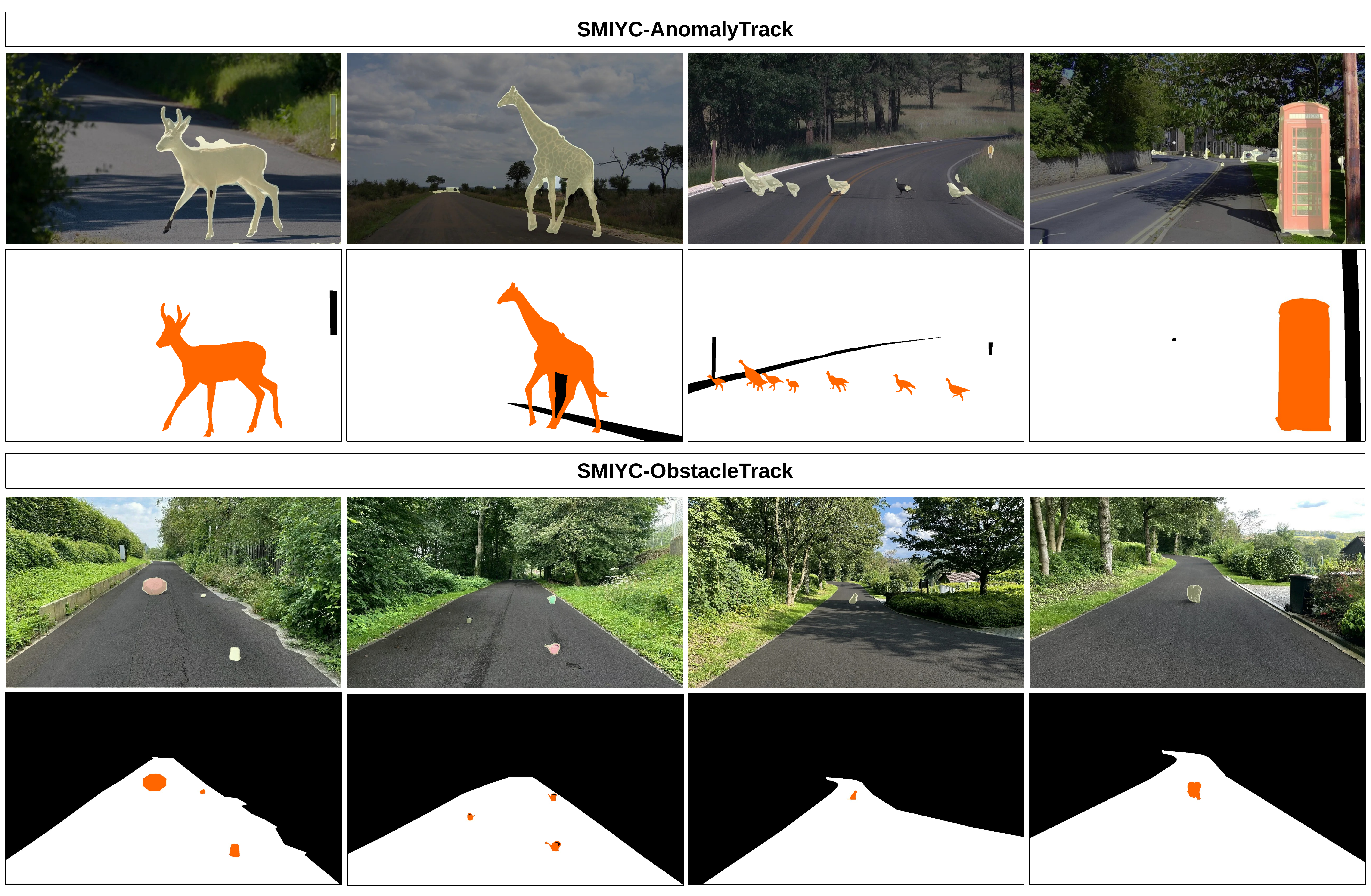}
    \caption{Anomaly detection performance of DenseHybrid on SMIYC validation subsets}
    \label{fig:smiyc_more}
\end{figure}

Figure \ref{fig:smiyc_more} visualizes anomaly detection performance of DenseHybrid on SMIYC-AnomalyTrack and SMIYC-ObstacleTrack.
Anomalies detected with DenseHybrid are highlighted above the input image.
The corresponding ground-truth has anomalies designated in orange. 
Inlier pixels are designated in white, while the ignore pixels are designated in black.

\end{document}